**Investigating U.S. Consumer Demand for Food Products with Innovative Transportation Certificates Based on Stated Preferences and Machine Learning Approaches**

Jingchen Bi[a], Rodrigo Mesa-Arango[a*].

[a] Department of Mechanical and Civil Engineering. Florida Institute of Technology, 150 W. University Blvd. Melbourne, FL 32901, U.S.A.

Email addresses: jbi2014@my.fit.edu (J. Bi), rmesaarango@fit.edu (R. Mesa-Arango).

* Corresponding author.

## ABSTRACT

This paper utilizes a machine learning model to estimate the consumer's behavior for food products with innovative transportation certificates in the U.S. Building on previous research that examined demand for food products with supply chain traceability using stated preference analysis, transportation factors were identified as significant in consumer food purchasing choices. Consequently, a second experiment was conducted to pinpoint the specific transportation attributes valued by consumers. A machine learning model was applied, and five innovative certificates related to transportation were proposed: Transportation Mode, Internet of Things (IOT), Safety measures, Energy Source, and Must Arrive By Dates (MABDs). The preference experiment also incorporated product-specific and decision-maker factors for control purposes. The findings reveal a notable inclination towards safety and energy certificates within the transportation domain of the U.S. food supply chain. Additionally, the study examined the influence of price, product type, certificates, and decision-maker factors on purchasing choices. Ultimately, the study offers data-driven recommendations for improving food supply chain systems.

# 1. Introduction

In this research, we explored the U.S. market's interest in food products accompanied by transportation certificates. Employing a machine learning model, we utilize Stated Preference data to shed light on consumer preferences between products with and without transportation certificates. Our research is spurred by the increasing awareness among Americans concerning food supply chains and the capacity of transportation elements to address health considerations, improve quality of life, and promote fairness. This study significantly enhances the current understanding of food demand dynamics, offering actionable insights that are valuable for policymakers and industry stakeholders alike.

Data collection involved conducting a comprehensive online survey across the entire United States based on stated preferences (SP). Machine learning methodologies were employed in the subsequent analysis to extract meaningful insights. This research provides valuable insights for policymakers to develop effective food safety regulations, while industry stakeholders benefit from a better understanding of consumer preferences for informed decision-making. Additionally, it contributes to agricultural and transportation economics by filling knowledge gaps on the demand for traceable food products in the U.S. market and sets the stage for future research endeavors.

FSCT is crucial for risk mitigation, preventing the distribution of unsafe products and averting legal liabilities, negative public perceptions, and costly product recalls. Food safety measures encompass various stages of the supply chain, from production to marketing, aiming to shield consumers from potential hazards. It is essential to ensure consumer awareness aligns with regulations mandating traceability throughout production and distribution. There are established food safety certification systems within the supply chain that provide detailed information on the handling and processing of food products under certain regulations in different regions, such as ISO 22000- International Organization for Standardization 22000 (Bomba & Susol, 2020; Carmen Paunescu , Ruxandra Argatu, Miruna Lungu, 2018), SQF- Safe Quality Food (Bomba & Susol, 2020; Mohammed & Zheng, 2017; Schuster & Maertens, 2015; Seok et al., 2016), IFS- International Featured Standards (Bomba & Susol, 2020; Ehrich & Mangelsdorf, 2018; Latouche & Chevassus-Lozza, 2015; Schuster & Maertens, 2015), FSDC 22000- Food Safety System Certification 22000 (Bomba & Susol, 2020; Mohammed & Zheng, 2017), Global G.A.P.- Global Good Agricultural Practices (Bomba & Susol, 2020; Mohammed & Zheng, 2017; Seok et al., 2016), BRC- Brand Reputation Compliance (Bomba & Susol, 2020; Latouche & Chevassus-Lozza, 2015; Mohammed & Zheng, 2017; Schuster & Maertens, 2015), and HACCP- Hazard Analysis and Critical Control Points (Kafetzopoulos et al., 2013; F. Liu et al., 2021). This study's implications resonate deeply within the food supply chain, consumer behavior, and sociology domains, particularly given recent events highlighting the importance of transportation and its impact on consumers.

To gain a deeper insight into how transportation influences the food supply chain, it is crucial to examine the role of recent technological progress in facilitating the integration of FSCT.

Existing research on the demand for FSCT has primarily focused on Asian and European regions. However, despite growing interest in FSCT, there are significant gaps in the literature that require further investigation. While some studies have explored FSCT in the U.S., there is a lack of comprehensive research examining this demand from diverse perspectives and evaluating preferences for various traceable food products. Additionally, the implications of tracing each key element within food supply chains on consumer food product choices have not been fully explored. Furthermore, the interplay between FSCT, price, food categories, certifications, and consumer behavior remain unclear. Therefore, conducting additional research is essential to bridge these gaps and achieve a more comprehensive understanding of the subject. All these gaps were addressed by the discrete choice model experiment we presented in a previous study (Bi & Mesa-Arango, 2024). The findings indicate that transportation significantly influences consumers' food purchase preferences. Consequently, we narrow our focus to transportation aspects to identify specific factors that consumers prioritize.

In our second experiment, we used machine learning model instead of discrete choice model for the following reasons:

*Flexibility*: Machine learning classification models, such as decision trees, random forests, or support vector machines, are highly flexible and can capture complex patterns in the data more effectively than discrete choice models. They can handle nonlinear relationships between predictors and outcomes, allowing for more accurate predictions.

*Feature Engineering*: Machine learning classification models can automatically handle feature engineering, extracting relevant features from the data without requiring manual specification of utility functions or choice set construction, as is often the case with discrete choice models.

*Handling Imbalanced Data*: Stated preference data may sometimes be imbalanced, with one class significantly outnumbering the others. Machine learning classification models can handle imbalanced datasets more effectively through techniques such as class weighting, resampling, or using algorithms designed to handle class imbalance.

*Scalability*: Machine learning classification models are often more scalable and can handle larger datasets with higher dimensionality compared to discrete choice models. This scalability allows for more comprehensive analysis of stated preference data.

*Ensemble Methods*: Machine learning classification models often utilize ensemble methods, such as bagging or boosting, which combine multiple models to improve predictive performance. These ensemble methods can further enhance the accuracy and robustness of predictions compared to discrete choice models.

*Generalization*: Machine learning classification models can generalize well to unseen data, capturing underlying patterns and relationships that may not be explicitly specified in the model. This ability to generalize is particularly useful for stated preference data, where the goal is often to predict preferences for new scenarios or products.

While discrete choice models have been widely used for stated preference data, machine learning classification models offer several advantages which prompted to be employed in the second experiment.

In our second study, five innovative transportation certificates were proposed: transportation mode, energy source, IoT, safety, and MABDs certificate. Results show a clear preference for products with transportation certificates, particularly safety and energy source certificates. Pricing and various food certificates significantly influence purchasing decisions. Sociodemographic factors including age, gender, educational level, type of employment, household type and income, and regional location also play pivotal roles in influencing consumers' purchasing decisions within the food market.

Hence, this study offers five significant contributions within the realm of transportation economics research: (i) It provides valuable insights into the demand for food products with transportation certificates in the United States; (ii) The study simultaneously analyzes multiple categories of food products, broadening the scope of investigation; (iii) It assesses the influence five transportation certificates on the demand for food products; (iv) The study elucidates the intricate balance between factors such as food product pricing, certifications, transportation traceability, and decision-maker considerations; (v) By developing a comprehensive framework that articulates stated preferences, machine learning, and explained artificial intelligence for analyzing food product purchases involving transportation certificates, this work is expected to enhance understanding in this field and contribute to the formulation of effective policies and strategies for managing food supply chains.

The paper provides a well-organized framework for investigating the demand for food products incorporating transportation certificates. Its structure is as follows: In the introductory section, the paper offers a succinct yet comprehensive introduction and rationale for the study. The second section delves into a thorough examination of the current literature on machine learning method in food supply chain, emphasizing the prevailing trends and identifying research gaps within the field. Section 3 presents the research methodology, offering a detailed account of the proposed approach. Following this, in Section 4, the paper applies the methodology to a specific case study, showcasing its practicality and relevance. In Section 5, the results of the study are laid out, accompanied by recommendations that are drawn from the findings. Finally, Section 6 draws the paper to a conclusion. It discusses the implications of the research and outlines potential avenues for future investigations within this domain.

## 2. Literature Review

The following literature review machine learning's applications in food supply chain among different regions, including China, India, the United Kingdom, the United States, Spain, and beyond. The literature emphasizes that machine learning has found application various domains within the food industry, including food sorting, packaging, distribution, demand prediction, and consumer behavior studies.

Existing studies often suffer from regional lack consideration of factors such as conjoint effects, food types, and decision-maker factors that influence consumer behavior. Moreover, the literature has not adequately addressed the specific impacts of various transportation elements in food supply chain and the complex trade-offs consumers make between FSCT, price, food type, certifications, and other behavioral factors. Research gaps in this domain motivate the present investigation to fill these voids by introducing a new methodology and case study to delve deeper into these issues.

*2.1. Research in Food Sorting*

Food sorting plays a critical role in optimizing food quality, safety, and efficiency throughout the supply chain, ultimately benefiting producers, retailers, and consumers alike. By removing damaged, defective, or contaminated items, food sorting enhances product quality and customer satisfaction. Machine learning techniques can be employed for various purposes in food supply chain soring process, including detecting food types, classifying damage, and assessing the ripeness of food products.

Machine learning has been utilized for food classification purposes, such as fruit and vegetables (Bhargava et al., 2022; Zou et al., 2022), walnut (Magnus et al., 2021) and adulteration classification of various substances such as cereal (Bai et al., 2022).

Bhargava et al. (2022) used Linear Regression (LR), sparse representation-based classification (SRC), Artificial Neural Network (ANN), and Support Vector Machine (SVM) to detect vegetables (jalapeno, lemon, sweet potato, cabbage, and tomato) and fruits (apple, avocado, banana, and orange) types in India. The algorithm achieves over 85% for detection of type.

In the research of Zou et al. (2022), CatBoost, XGBoost, SVM, and K-Nearest Neighbors (KNN) machine learning algorithm were utilized to classify apple's colors in China. The model successfully classified apple samples into three types (red, light red, and light yellow), achieving a recognition rate of 96.7%.

Magnus et al. (2021)combined Linear Discriminant Analysis (LDA), Quadratic Discriminant Analysis (QDA), SVM, Extreme Learning Machine (ELM) and Partial Least Squares Discriminant Analysis (PLS-DA) to identify walnut from foreign objects and mold Belgium, showing a correct classification rate exceeding 98%.

In the study of Bai et al. (2022), a Back-Propagation Neural Network (BPNN) model was built to detect cereal-crop adulteration in China. The finding show that the model achieved an overall identification accuracy of 90%.

Machine learning has also been employed to identify damage in food products, such as fruits (apples and mangos) (Hemamalini et al., 2022), cookie (Drogalis et al., 2024), and unwashed eggs (Nasiri et al., 2020).

Hemamalini et al. (2022) used KNN, SVM, and C4.5 decision tree algorithm to classify fruits as damaged or good in India. The results show accuracy of 68.1% (C4.5), 84% (KNN), and 98% (SVM).

In the paper of Drogalis et al. (2024) a closed-loop system with ANN, machine vision, and robotics was created to detect the quality of cookies in the U.S., resulting in an efficient food quality inspection with a success rate of 98%.

Nasiri et al. (2020) applied a Convolutional Neural Network (CNN) to categorize unwashed eggs in Iran into three classes: intact, bloody, and broken. achieving an average overall accuracy of 94.84%.

Machine learning has also found application in differentiate the ripeness of food product, such as dragon fruit (Patil et al., 2021), banana (Chu et al., 2022; Xie et al., 2018), and nectarine (Munera et al., 2017).

Patil et al. (2021) proposed a working process to classify the maturity level of dragon fruit in India using CNN, ANN, and SVM models. Xie et al. (2018) and Chu et al. (2022) both applied PLS-DA to detect ripeness of bananas in China, obtaining a good calcification result (over 91.53% in validation set). Munera et al., (2017) utilized PLS model to detect the ripeness of nectarine in Spain, and optimal results were achieved with R2 values exceeding 0.87.

*2.2. Research in Food Packaging*

Food packaging plays a vital role in ensuring the safety, quality, convenience, and sustainability of food products, benefiting both consumers and the food industry. Machine learning algorithms can analyze images of food packaging to detect defects, such as tears, wrinkles, or misalignments, ensuring that only products meeting quality standards are sent to market. Machine learning has been found various applications in food packaging, including quality inspection (Banús Paradell et al., 2021) and chemical migration (Medus et al., 2021; S.-S. Wang et al., 2023).

S.-S. Wang et al. (2023) investigated chemical migration from food packaging materials to food (water, milk, wine, and aqueous ethanol) in Taiwan using CatBoost, LightGBM, LightGBMXT, LightGBMLarge, RF (random forest), ExtraTrees, KNNDist, KNNUnif and WE (weighted ensemble). The results revealed that all generic models outperformed the existing model, with an average R2 of 0.860 for test performance.

In the research of Medus et al. (2021), the utilization of CNN was introduced as a classifier for heat-sealed food packages in the production line in Spain. The outcomes demonstrate that the global fault detection accuracy for harmful impurities exceeded 94%.

Banús Paradell et al. (2021) introduced a computer vision system utilizing the CNN algorithm to automate the sealing and tightness inspection of pizza packages in U.S. The evaluation revealed that the system achieves optimal results, with mean precision values reaching 99.87%.

*2.3. Research in Food Distribution*

The distribution and retail phase serves as the link between food production and processing and the final consumption, thus completing the farm-to-fork cycle (Manzini et al., 2019). Machine Learning (ML) can help decide delivery routes, predict food demand, supply raw materials, and plan logistics, thus offering several applications in food distribution, including transportation (Buelvas Padilla et al., 2018; Q. Liu & Xu, 2008; Rabbani et al., 2016; X. P. Wang et al., 2018) and storage (Emsley et al., 2022; Kumar S.V. et al., 2020; Ropelewska & Noutfia, 2024; Siripatrawan & Makino, 2018; Sricharoonratana et al., 2021).

Machine learning is increasingly being applied in various aspects of food transportation to improve efficiency, safety, and sustainability. Machine learning algorithms analyze historical traffic data, weather patterns, road conditions, and delivery schedules to optimize delivery routes. This helps reduce transportation costs, fuel consumption, and delivery times while improving overall efficiency.

Liu & Xu (2008) studied optimization of perishable products delivery in China with Variable Neighborhood Search (VNS) and a genetic algorithm (GA). The algorithm has been shown to significantly reduce computation time and greatly enhance the quality of solutions.

In the paper of X. P. Wang et al. (2018), vehicle routing problem in China regarding the delivery of fresh agricultural products was investigated using GA and discovered that the proposed model serves as an effective tool for decision-makers to make delivery decisions efficiently, especially under random fuzzy environments.

Rabbani et al. (2016) also explored vehicle routing problem in Iran for perishable food delivery utilizing GA. The results from the model have shown promise, particularly in terms of CPU time efficiency for addressing large-scale problems when compared to exact solvers.

In their study, Buelvas Padilla et al. (2018) analyzed vehicle routing problem in Colombia to minimize perishable food damage, taking into account road conditions employing GA. The results were able to show the connection between damaged products and distance traveled, demonstrating that the proposed solution approaches can generate feasible routes while balancing these objectives.

Machine learning studies in food storage focus on leveraging data-driven approaches to optimize storage conditions, ensure food safety, minimize waste, and enhance operational efficiency.

Kumar S.V. et al. (2020) developed an autonomous food wastage control system in India using Neural Network Model. The results demonstrated a minimum training error (mean squared error) for expiry date prediction of 0.4268.

In the paper of Ropelewska & Noutfia (2024), the effects of storage conditions on grape's behavior in Poland was evaluated employing Bayes Net, Multilayer Perceptron, Kstar, and Random Forest algorithm. The results successful classified the grapes into three categories (grapes stored in the freezer, in the refrigerator and in the room) with an overall accuracy of 96%.

Emsley et al. (2022) created a model with SVM to predict the impact of storage conditions on tomatoes in U.K. The model was proved to successfully estimate the time-after-harvest despite the storage condition with classification accuracy across exocarp (92%) and locular gel (84%) samples.

In the research of Siripatrawan & Makino (2018), the quality attributes and shelf life of packaged sausage in Thailand was assessed using Partial Least Squares Regression (PLSR) and ANN. The prediction model was confirmed as a reliable tool for assessing the quality characteristics of the packaged sausages.

Sricharoonratana et al. (2021) proposed a model to determine shelf life of cakes in Thailand using PLSR and PLS-DA and the findings revealed a prediction accuracy of 91.3%.

*2.4. Research In Forecasting/ Demand Prediction*

Accurate food demand prediction helps prevent issues such as overstocking, overproduction, and excessive resource utilization (Hofmann & Rutschmann, 2018). By improving the accuracy of demand forecasts, machine learning contributes to better inventory management, reduced wastage, optimized production schedules, and enhanced customer satisfaction. The use of ML algorithms helps in improving demand forecasting and production planning (Feng & Shanthikumar, 2018).

Veiga et al. (2016) performed a demand forecasting for perishable food in Brazil based Wavelets Neural Networks (WNN). The finding indicates that the proposed model achieved a demand satisfaction rate exceeding 98% across all three categories of perishable food products.

In their study, Lutoslawski et al. (2021) studied food demand prediction in Poland with the employment of the Nonlinear Autoregressive Exogenous Neural Network (NARXNN). In general, the proposed model achieved satisfactory predictive outcomes.

In the research of Jayapal (2023), food demand prediction for food orders form an Italian restaurant in Ireland was investigated using Multiple Linear Regression (MLR), Lasso and Ridge Regression (LRR), Bayesian Ridge Regression (BRR), Support Vector Regressor (SVR), Decision Tree Regression (DTR), Random Forest Regression (RFR), Gradient Boosting Regression (GBR), XGBoost Regression, CatBoost Regression, Facebook prophet, and HyperParameter Tuning. The result reveals that XGBoost and LightGBM The findings indicate that XGBoost and LightGBM surpassed all existing boosting algorithms, achieving a Root Mean Squared Error (RMSE) of 0.45 and 0.49, respectively.

Nosratabadi et al. (2021) predicted Production for livestock yield, live animals, and animal slaughtered in Iran using Multilayer perceptron (MLP) and Adaptive Network-based Fuzzy Inference System (ANFIS). The findings disclosed that the ANFIS model exhibited the lowest error rate in predicting food production.

In the paper of Kantasa-ard et al. (2021), the demand forecasting for agricultural products in Thailand evaluated adopting Long Short-Term Memory (LSTM) method. The observation tells that holding cost varied by approximately 0.09–1% between forecast and real demand, and the transportation cost varied from 0.3–1.07%.

Chelliah et al. (2024) analyzed demand forecasting of agriculture in India using CNN algorithm. The study presented an enhanced accuracy of up to 96 percent.

*2.5. Research In Consumer Behavior*

Across various organizations, it has been observed that AI is not only aiding food processing industries in creating diverse flavor combinations but also assisting customers in selecting innovative tastes (Dehghan-Dehnavi et al., 2020). Machine learning techniques find applications in the food retailing phase to forecast consumer demand, understand perception, and analyze buying behavior (De Sousa Ribeiro et al., 2018). Machine learning enables businesses in the food supply chain to gain valuable insights into consumer behavior, improve customer satisfaction, and drive business growth through targeted marketing, personalized experiences, and optimized supply chain management.

Borimnejad & Eshraghi Samani (2016) analyzed consumer's behavior for packed vegetable in Iran using ANN. The results revealed that demographic factors, alongside traditional economic variables like price and income, play a significant role in influencing customer choices.

In their research, Nilashi et al. (2021) evaluated customer's preferences in vegetarian restaurants in Malaysia employed Latent Dirichlet Allocation (LDA), Self-Organizing Map (SOM), Classification and Regression Tree (CART). The finding indicates that the predictive accuracy of the proposed method surpasses that of other methods developed solely using supervised learning techniques.

Shen et al. (2021) assessed consumer's buy and pay preferences for labeled food products in U.S. with Logistic Regression (LR), SVM, RF, and NN. In general, the four machine learning models exhibited an average performance surpassing that of random guessing, which typically achieves a prediction accuracy of 0.33.

In the study of Naik (2020), an intelligent food recommendation system in India was designed using GA. Based on the flavor preference of respondents, food product is recommended to the purchasers likeliness.

Fiore et al. (2017) predicted consumer's healthy choices regarding type 1 wheat flour in Italy with the application of NN, RF, and SVM. The outcome suggested that consumer is willing to pay higher price for "type 1" wheat flour rather than basic types of wheat flour due to health concerns.

In the paper of Lilavanichakul et al. (2018), consumer's purchase demand for imported ready-to-eat foods in China was classified utilizing ANN, MLP, and Backward Stepwise Logistic Regression Model. The results indicate that the backward stepwise logistic regression model classified the consumer purchasing decision better than the MLP model with a correct classification of 78.8% for Beijing and 74.6% for Shanghai.

*2.6. Research Gap*

This comprehensive literature review comprises a series of research papers investigating machine learning's applications in food supply chain among different regions, including China, India, the United Kingdom, the United States, Spain, and beyond. The literature emphasizes that machine learning has found application various domains within the food industry, including food sorting, packaging, distribution, demand prediction, and consumer behavior studies.

However, this review has identified several significant research gaps within this domain. These gaps include unexplored consequences of specific transportation elements on consumer choices, the absence of studies addressing consumer behavior in the United States, a limited focus on diverse food product types with FSCT, and a dearth of investigations into the intricate trade-offs consumers make between FSCT, price, food type, certifications, and other behavioral determinants.

In summary, there exist gaps in the literature concerning the analysis of consumer preferences for food products with transportation certificates in the U.S. using machine learning modeling methods. This investigation aims to address this gap by employing the methodology and case study presented in the following sections.

## 3. Methodology

Previous studies have proposed methods for tracing food supply chains, yet none have delved into understanding the specific preferences of U.S. consumers regarding food products with transportation certificates using machine learning techniques. Machine learning empowers precise forecasts of forthcoming trends, aiding in planning and decision-making processes. This paper examines the process of choosing between product alternatives using a novel methodology that integrates stated preferences, commonly utilized in econometric research (Mesa-Arango et al., 2021; Mesa-Arango & Ukkusuri, 2014; Nair & Mesa-Arango, 2022), with machine learning classification techniques (Pineda-Jaramillo, 2021; Pineda-Jaramillo et al., 2021, 2022; Pineda-Jaramillo & Arbeláez-Arenas, 2021, 2022).

**Figure 1** presents the 5-step method that combines machine learning and stated preferences to produce valuable behavioral insights on the phenomenon under scrutiny. The approach consists of a series of steps, each of which is discussed in detail in the corresponding subsection below.

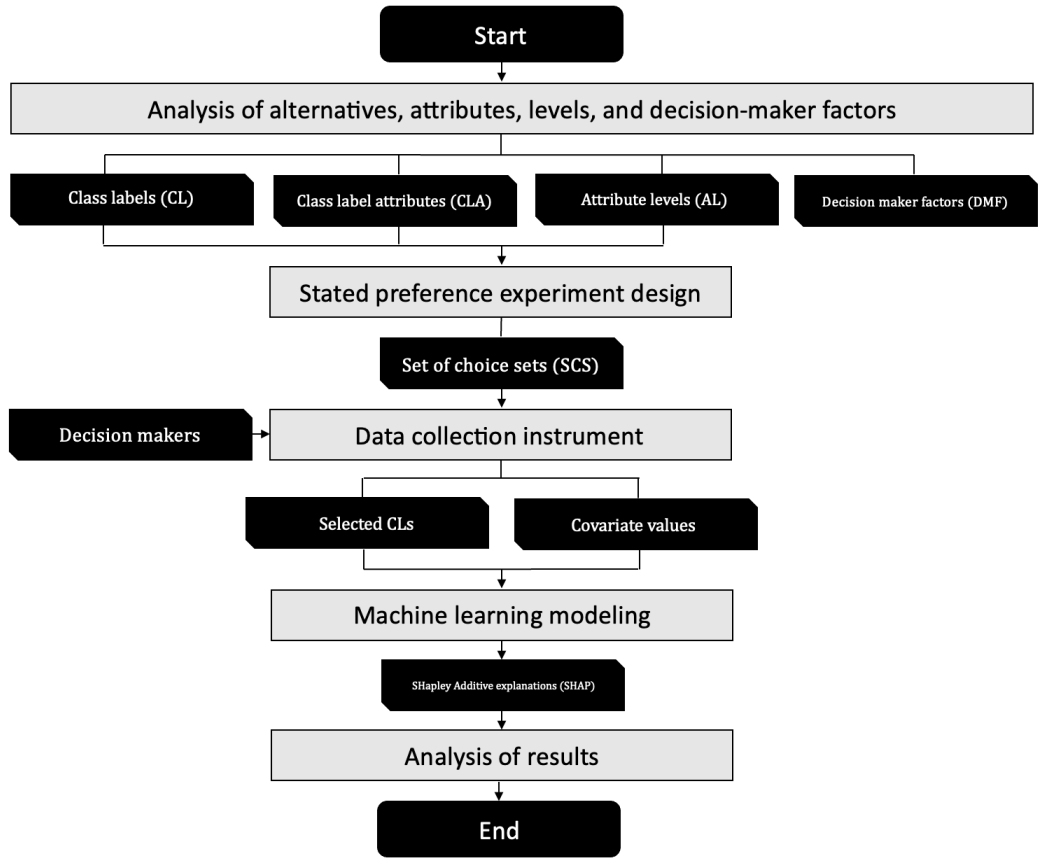

**Figure 1**. Method to generate demand-driven management guidance from stated preferences with discrete-choice modeling.

*3.1. Analysis of Alternatives, Attributes, Levels, And Decision Maker Factors*

In the typical discrete decision-making process (**Figure 2**), the decision-maker faces the task of choosing a particular label from a choice set containing all available alternatives. In the context of stated preferences, this set encompasses both real and hypothetical alternatives or class labels.

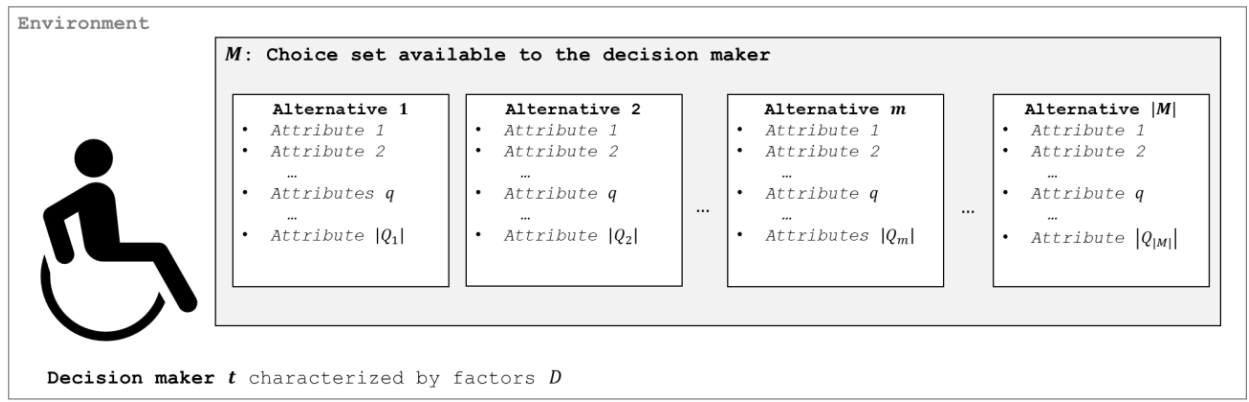

**Figure 2**. Illustration of the decision-making process in the context of discrete choices.

The decision-maker's preference for each class label is influenced by the utility it offers and a series of class label attributes. This utility is shaped by a range of decision-maker factors. Environmental factors that affect alternative selection are considered part of the class label attributes, while those that do not impact selection are categorized as decision-maker factors. In stated-preference experiments, each class label attribute is defined by a range of attribute levels.

In conceptualizing, class labels, class label attributes, attribute levels, and decision-maker factors are identified through direct observation of the decision-making process, expert insights, focus groups, or comprehensive literature reviews.

There are five recommendations for designing each attribute level to enhance data collection size and model quality (Ke et al., 2017a). These include:

(i) Maximizing information while minimizing attribute level size.

(ii) Reducing the least common multiple among attribute-level sizes.

(iii) Ensuring that extreme values of attribute levels span a wide range (Bliemer & Rose, 2005).

(iv) Ensuring that attribute level values are logical and credible.

(v) Including at least 2 attribute levels per class label attribute.

*3.2. Stated Preferences Experiment Design*

Once the definitions of class labels, class label attributes, and attribute levels are established, they are utilized to create a series of choice sets for data collection. This procedure, known as stated preferences experiment design, involves determining the number of choice tasks each respondent undertakes and the combination of attribute levels for each attribute in each class label within each choice set (Louviere & Hensher, 1983; Louviere & Woodworth, 1983; Ortúzar & Willumsen, 2011; Reed Johnson et al., 2013; Walker et al., 2018).

The set of choice sets is depicted as a matrix (**Figure 3**), where each row represents a choice set. If class labels and class label attributes are labeled with positive ordinal numbers, each column denotes the $n^{th}$ attribute of the $m^{th}$ class label. Therefore, a stated preferences experiment design specifies the attribute level value that should be presented to a respondent analyzing the corresponding choice set scenario.

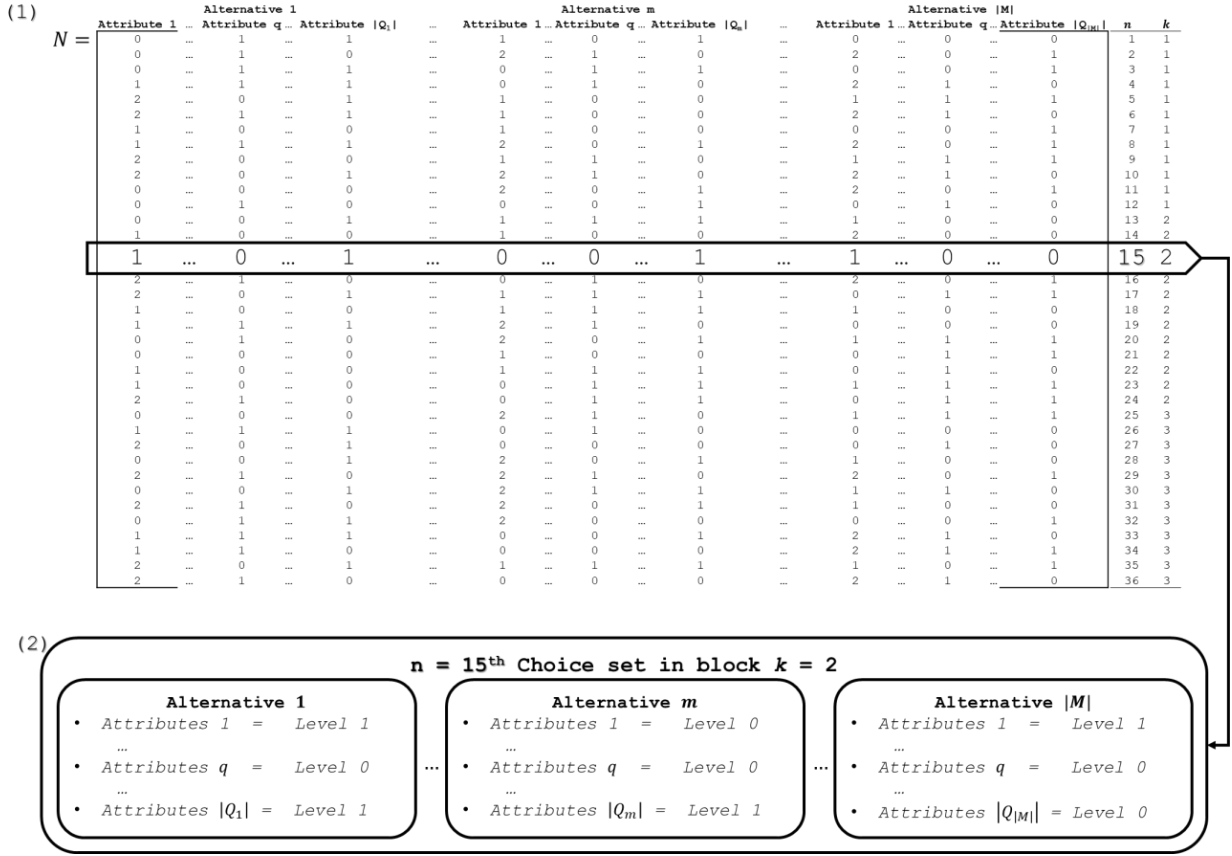

**Figure 3**. Converting a numerical choice set $n \in N$ into the scenario for the n[th] stated preference survey – an illustrative example.

*3.3. Data Collection Instrument*

The decision-maker factors and the set of choice sets are integrated into a four-module data collection instrument (**Figure 4**), which solicits inputs from a group of decision-makers and consolidates their responses.

In the survey introduction (Module 1), respondents are introduced to the overall data collection process. This module clarifies elements regarding privacy and participation, confirms respondent eligibility and consent, and prepares them for the subsequent data collection modules.

Module 2 of the decision-maker data collection encodes the decision-maker factors to solicit respondent entries, ensuring proper storage of the collected data.

Module 3 of the stated preferences data collection involves converting the numerical set of choice sets into intuitive questions for respondents (**Figure 4**). Initially, the module introduces the stated preferences experiment to the respondent. Subsequently, each respondent is assigned to a block, and they evaluate each scenario within the corresponding set of choice sets, selecting the most desirable class label described in each case.

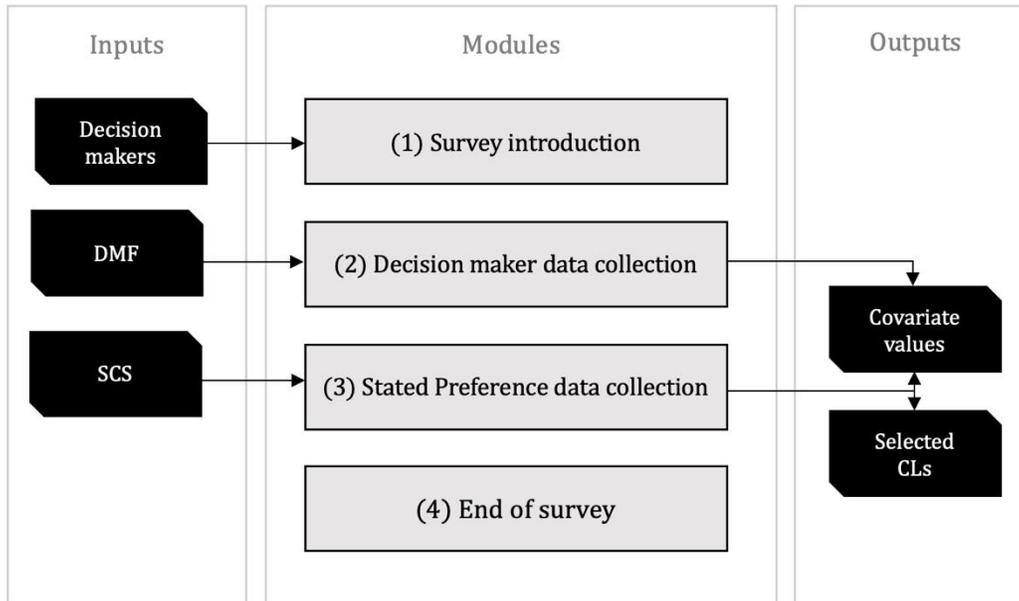

**Figure 4**. Data collection instrument: inputs, modules, and outputs.

Concluding the survey, Module 4 formally ends the process by expressing gratitude to respondents, gathering feedback for ongoing enhancements, and providing details on subsequent steps such as payment claims.

*3.4. Machine Learning Modeling*

Machine learning models are emerging as a compelling and favored substitute for traditional random utility models, which have conventionally dominated food purchase preference analysis in transportation and behavioral research. These models automate data-driven algorithms and systematically detect non-linear relationships within datasets, a capability unavailable in typical random utility models. Among the most popular machine learning models, such as decision trees, random forest, gradient boosting, and artificial neural networks, they have found widespread application in transportation research, offering enhanced reliability and accuracy compared to random utility models. (Hagenauer & Helbich, 2017; Pineda-Jaramillo & Arbeláez-Arenas, 2022; F. Wang & Ross, 2018).

Therefore, machine learning modeling constructs a classification model to forecast the analyzed class label and assess its classification efficacy based on attributes of the class label and decision-maker factors.

Initially, the gathered dataset is partitioned into a *training set*, utilized to train the prediction models, and a *test set*, employed to assess the models' performance.

As this research encompasses multiple class labels, five supervised machine learning classification algorithms are utilized, based on their demonstrated strong multiclass prediction performance in prior studies (Hagenauer & Helbich, 2017; Ke et al., 2017b; S. M. Lundberg et al., 2018; Pineda-Jaramillo et al., 2021, 2022, 2024; Rashidi et al., 2021; Servos et al., 2020; Zhao et al., 2020). These models consist of: (i) gradient boosting classifier (implemented with the LightGBM framework), (ii) extra trees classifier, (iii) random forest classifier, (iv) k neighbors classifier, and (v) decision tree classifier. They possess the capability to manage high-dimensional data effectively and deliver precise outcomes. Subsequent sections provide detailed insights into each model.

The gradient boosting classifier, implemented with the LightGBM framework, is an ensemble learning approach that enhances upon the decision tree algorithm. It employs a boosting technique to progressively refine the performance of decision trees by concentrating on the data points misclassified in previous iterations. Renowned for its high accuracy and proficiency in managing large and intricate datasets, gradient boosting does demand significant computational resources and meticulous hyperparameter tuning.

The extra trees classifier, alternatively termed extremely randomized trees, is an ensemble learning technique that extends the decision tree algorithm. It randomly chooses subsets of the data and features to construct numerous decision trees, subsequently amalgamating their predictions to formulate a final prediction. While this approach can enhance the accuracy and robustness of the decision tree model, it may also necessitate increased computational resources.

The random forest classifier, also an ensemble learning method rooted in decision trees, constructs numerous decision trees utilizing subsets of the data and features. Subsequently, it amalgamates their predictions to formulate a final prediction. Renowned for its proficiency in managing large and intricate datasets, as well as its high accuracy and robustness, the random forest classifier may present challenges in interpreting its decision-making process.

The k neighbors classifier operates on the principle that similar data points usually share similar labels or values. During training, it stores the entire dataset for reference. When predicting, it calculates distances between the input data point and all training examples using a selected metric like Euclidean distance. Then, it identifies the K nearest neighbors based on these distances. For classification, it assigns the most common class label among the K neighbors to predict the label for the input.

The decision tree classifier, a supervised learning algorithm frequently applied to classification tasks, originates from decision analysis and relies on tree structures. Renowned for their interpretability, decision trees offer a straightforward representation of decision-making processes. Nonetheless, they may suffer from overfitting, particularly when dealing with larger datasets.

Every machine learning model possesses distinct hyperparameters that can be fine-tuned to enhance its predictive performance. Hyperparameter tuning typically employs a random search method (Bergstra & Bengio, n.d.) to assess parameter values with significant effects on each model's performance. The area under the receiver operating characteristics curve (AUC), recognized for its objectivity in evaluating multiclass classification models, serves as the metric for optimization performance measurement.

Equation (1) calculates the $AUC$ by considering the true positives $TP$, representing correct predictions of the positive class, true negatives $TN$, indicating accurate predictions of the other classes, and a dummy variable $\delta_{ij}$ dependent on the probability scores $p_i$ and $p_j$ assigned by the model to points $i$ and $j$ within the area-under-the-receiver-operating-characteristics curve (as described in Equation (2)).

$$AUC = \frac{1}{(TP)(TN)} \sum_{i=1}^{TP} \sum_{j=1}^{TN} \delta_{ij} \qquad (1)$$

$$\delta_{ij} = \begin{cases} 1 & if\ p_i > p_j \\ 0 & otherwise \end{cases} \qquad (2)$$

The stratified *k*-fold cross-validation method is employed to validate trained machine learning models. It involves randomly partitioning the training set *k* groups, ensuring each group maintains a similar composition concerning the various classes the model aims to predict. Subsequently, the machine learning model to validate is applied to *k-1* subgroups, while the remaining subgroup assesses model performance using the $AUC$ metric (Witten et al., 2016). This process iterates *k* times and the average $AUC$ is utilized to validate model performance. By doing so, this technique helps mitigate any bias that the machine learning model might introduce during its training phase.

*3.5. Analysis of Results*

The Shapley Additive explanation method (SHAP) (Witten et al., 2016) is employed to gauge the influence of class label attributes and decision-maker factors on the optimal machine learning model. This method delineates the direct effects and significance of individual class label attributes and decision-maker factors utilized in training the best machine learning model. Additionally, it estimates the degree of impact that each feature holds on class label classification.

SHAP computes the average discrepancies in predictions across all possible feature orderings, utilizing this precise foundation to generate SHAP values (S. Lundberg & Lee, 2017; Parsa et al., 2020) .This approach enables the extraction of the impact value and direct effect associated with each feature in classifying the predefined class label.

## 4. Case Study and Data

A practical case study that employs the previously outlined methodology is presented to gain insights into the U.S. demand for food product with transportation certificates within the framework of discrete decision-making. The analysis covers alternatives, attributes, levels, and decision-maker factors. It investigates the factors influencing choices related to food products, such as product types, price, certifications, transportation modes, safety certificates, energy source, and more. The study utilizes an SP experiment design to collect data, and the section provides details on the data collection instrument and summary statistics. The data offers valuable sociodemographic insights into respondents, their awareness of food supply chains, and typical food product purchases, shedding light on consumer behaviors and preferences.

*4.1. Analysis of Alternatives, Attributes, Levels, and Decision-Maker Factors*

This paper investigates the discrete decision-making process of whether to purchase food products with transportation certificates. Substitutes to that choice are products without transportation certificates or not to purchase a product. To keep the consistency of the research, three class labels considered are still: Product A (Product with transportation certificates), Product B (Purchase product without transportation certificates), and Not Purchase.

Eleven class label attributes are selected to describe food product choices, among which six are the same as in the first experiment, i.e., product type, price, USDA organic certificate, gluten-free certificate, non-GMO certificate, other kind of certificates. New attributes for product with transportation certificates are: transportation mode, safety measures, energy Source, IoT integration, and MABDs. **Table 1** shows the levels associated with each class label attribute.

Food product types are designed the same as in the first experiment (**Table 1**(a)), where three levels of product types are: fruits or vegetables, meat or seafood, dairy or eggs, and other food products.

Two levels of prices (**Table 1**(b)) are still generated based on respondents' input: 20% of the entered regular price and 80% of the entered regular price.

Four types of certificates that remains are: USDA Organic (c), Gluten-Free (d), Non-GMO (e), and other kind of certifications (f).

Regarding transportation certificate features, products are described by five transportation related attributes, i.e., transportation mode (g), safety measures (h), energy source (i), IoT integration (j), and MABDs (k).

(g) Transportation mode certificate indicates the transportation mode with highest utilization during food product transit as one of the following options: air, road, rail, and water.

(h) Transportation safety certificate provides assurance that food products are safe during transit.

(i) Transportation energy source certificate describes the energy method used during food product transit by one of the following options: fossil fuel, green energy (electricity and other environmental-friendly energies like biofuel, ethanol, natural gas etc.), and not available.

(j) An IoT certificate ensures the transportation company utilizes IoT for real-time asset tracking and efficient food delivery through data optimization.

(k) MABDs certificate ensures food products to be delivered to the retailer's store, warehouse, or another facility on time.

The collected decision-maker factors $D$ include zip code, year of birth, gender, income, level of education, work type, household type, number of children, race, and citizenship, which align with similar data collected by the U.S. Census.

**Table 1** Class label attributes and attribute levels used in the case study.

| Attributes ($Q$) | Levels $\ell_q$ |
|---|---|
| (a) Product type. | - Fruits or vegetables. |
| | - Meat or seafood. |
| | - Dairy or eggs. |
| | - Other. |
| (b) Price. | - 20% of regular price ($). |
| | - 180% of regular price ($). |
| (c) USDA Organic Certified. | - No. |
| | - Yes. |
| (d) Gluten-Free Certified. | - No. |
| | - Yes. |
| (e) Non-GMO Certified. | - No. |

| Attributes ($Q$) | Levels $\ell_q$ |
|---|---|
| | - Yes. |
| (f) Other kind of certification available. | - No. |
| | - Yes. |
| (g) Transportation Mode. | - Air. |
| | - Road. |
| | - Rail. |
| | - Water |
| (h) Safety Certificates. | - No. |
| | - Yes. |
| (i) Energy Source. | - Fossil Fuel. |
| | - Green Energy. |
| | - Not Available. |
| (j) IoT Certificate. | - No. |
| | - Yes. |
| (k) Must Arrive by Dates. | - No. |
| | - Yes. |

(g) to (k) only for $m = 1$ (Purchase product with FSCT).

$m = 3$ (Not to purchase) is not related to any attribute.

### 4.2. Stated Preferences Experiment Design

In the following section, we explore the development of an orthogonal sequential design, a crucial component of our research methodology. This design provides essential insights into the relationships among key variables, laying the groundwork for the analyses and findings that follow.

An orthogonal sequential design is constructed using the software Ngene (ChoiceMetrics, 2018) and generates $|N| = 72$ choice sets which are split into $|K| = 18$ blocks to allow each respondent to consider 4 cases per survey.

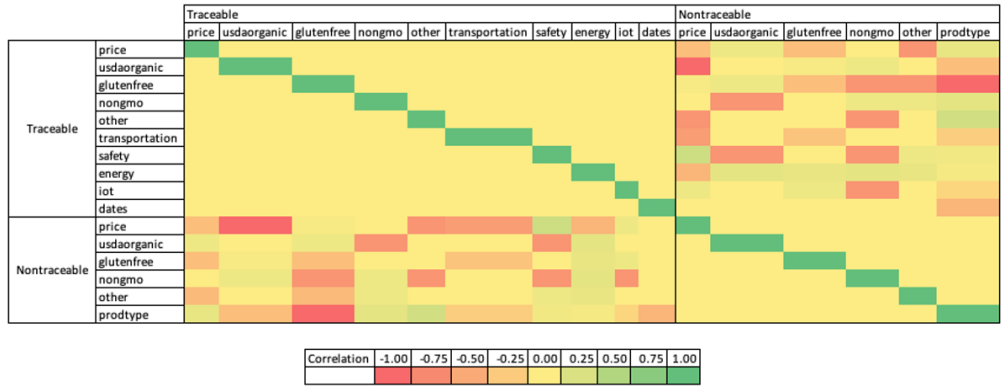

**Figure 5**. Correlation heatmap between variables in the stated preferences experiment.

**Figure 5** illustrates the low correlation between variables in the orthogonal $N$. Average correlation and standard deviation outside the diagonal are $0.46 \times 10^{-2}$ and 0.08, average correlation of variables within each alternative is 0.00 and correlation across alternatives ranges from -0.30 to 0.33.

Following the creation of an orthogonal SP experiment design, it is encoded along with the relevant information into the subsequent data collection instrument.

*4.3. Data Collection Instrument*

The previously constructed set of choice sets is embedded in an online survey implemented in Qualtrics (Qualtrics, 2021) and disseminated across the U.S. via Amazon Mechanical Turk (Amzon Mturk, 2021) from December 11 to 22, 2023. **Figure 6** exemplifies a SP choice set.

**Figure 6**. Example of a choice set presented to a respondent in the stated preferences experiment.

After data cleaning, 33,328 data points are collected and generated from 2083 respondents analyzing 3 alternatives in 4 scenarios. However, each respondent completed 4 rounds of 4-scenario questions, which make the total cases to be 16 in each survey.

**Figure 7** and **Figure 8** show the summary statistics of the data collected from the online survey. Generally speaking, most respondents are U.S. citizens, male, white, with a bachelor's degree, has 2 children, from a medium-income range, private workers, and from Midwest region.

72.0% of respondents have a bachelor's degree, 16.9% have a graduate or professional degree, 6.3% are high school graduates, 2.7% has associate degree, 1.6% went to some college, and 0.4% have less than high school degree.

Most respondents are private wage workers (50.2%), 39.2% are self-employed workers, 10.4% work for government, and 0.2% have unpaid work.

When talking about household annual income, 26.3% of the respondents earn $50K-$75K annually, 22.0% earn $75K-100K, 13% earns $35K-$50K, 12.5% earn 100K-150K, 9.1% earns 25K-35K, 6.3% earn $15K-25K, 5.7% earn 150K-200K, 2.9% earn 10K-15K, 1.2% earn less than 10K, and 1.0% earn more than 200K per year.

While 25.4% of respondents earn $50K-$75K annually, 20.1% earn $35K-$50K, 16.5% earn 75K-100K, 13.7% earn $25K-35K, 8.4% earn 15K-25K, 5.4% earn 10K-15K, 4.0% earn 100K-150K, 3.9% earn 150K-200K, 2.2% earn less than 10K, and 0.5% earn more than 200K per year.

Regarding children, 45.2% have two children, 28.3% have one child, 14.2% have three, 8% have no child, 4% have 4 children, and 0.3% have five or more children.

From household type, 49.8% belong to a couple family with children and 3.9% for couple family without children. 21.9% are male householder with children and 6.0 % are male householder without children, 13.0% are a female householder with children and 2.2% female householder without children, 3.1% are the nonfamily household (living alone), and 0.1% are other types.

Most of the respondents are U.S. citizens (99.7%). 64.2% have North American origin, 17.5% Latin American, 9.3% Asian, 5.8% European, 1.9% Africa, and 1.3% Oceanian origin.

88.6% of the respondents self-identify with the white race, 7.7% as Asian, 2.9% as black or African American, and 0.7% as other races.

36.4% of the respondents are from the Midwest U.S., 30.7% from the South, 23.3% from the West, and 9.6% from the Northeast.

65.0% of respondents are male and 35% are female.

The mean age of respondents is 38.55 years old while the minimum age is 21 years old and maximum age is 71 years old.

The survey asks respondents to identify a typical food product purchase. In the survey, respondents were asked to input a price for each product type from their last purchase. The price distribution of each category came from the data input.

Table **2** presents the price distributions for each category. Mean prices are $32.51 for fruit and vegetables, $38.11 for meat and seafood, $31.96 for dairy and eggs, and $32.79 for other food products.

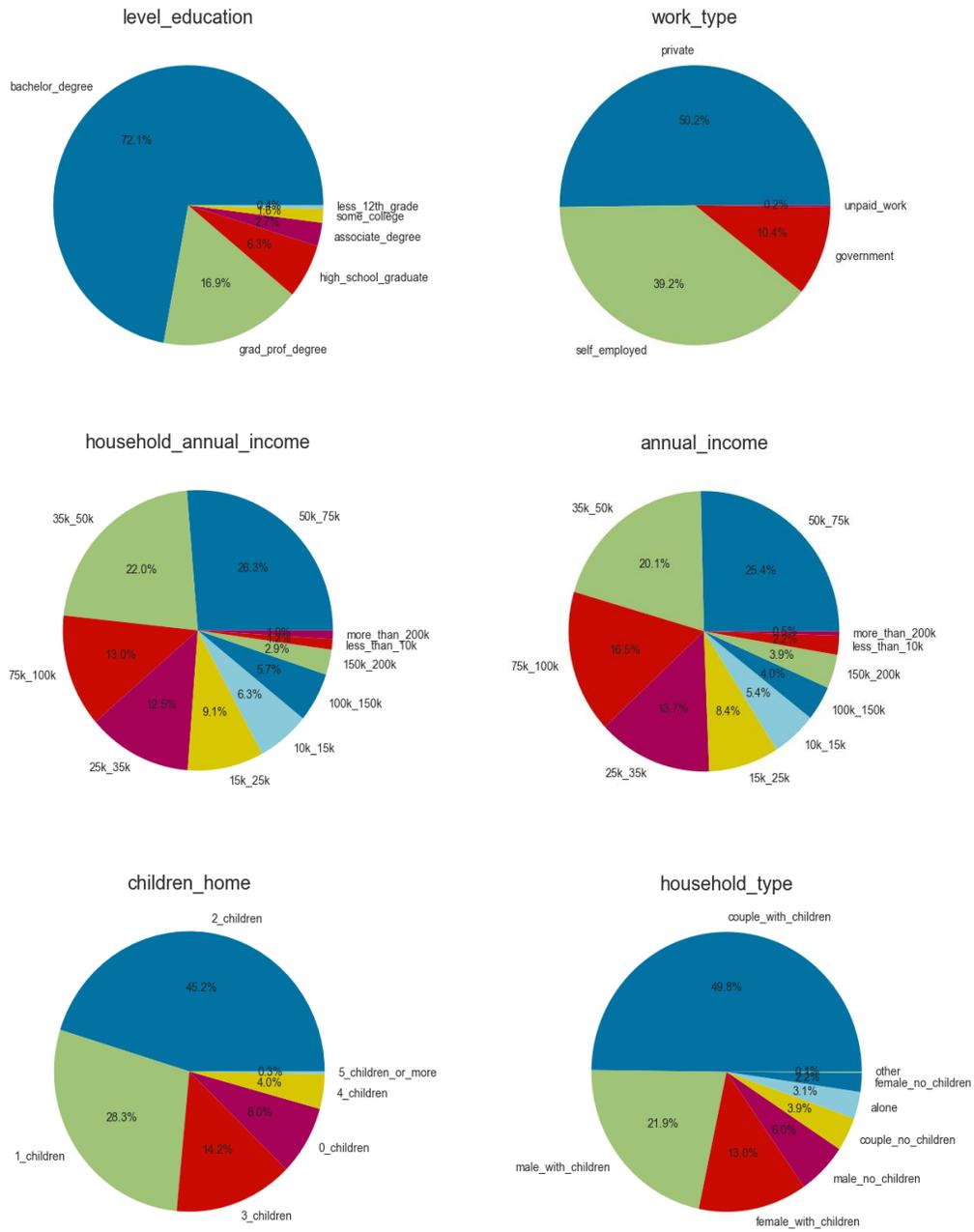

**Figure 7**. Summary statistics of selected decision maker-factors: (a) level of education, (b) work type, (c) household annual income, (d) annual income, (e) number of children at home, and (f) household type.

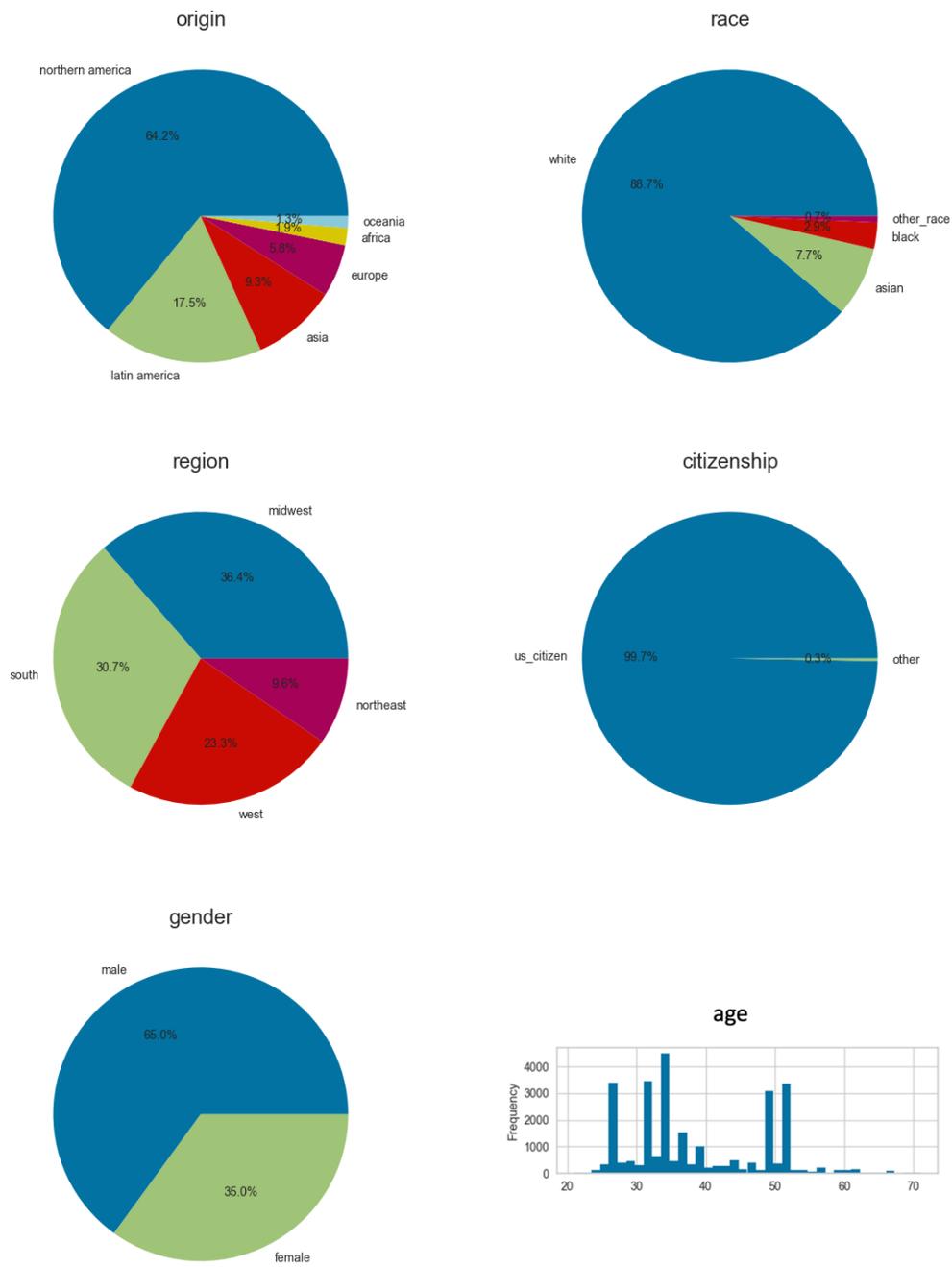

**Figure 8**. Summary statistics of selected decision maker-factors: (a) origin, (b) race, (c) region, (d) citizenship, (e) gender, and (f) age.

**Table 2** Statistic Summary: Prices for different food types.

| Variables | Mean | Min. | Max. |
|---|---|---|---|
| Price inputted by respondent for last purchased fruit and vegetable ($) | 32.51 | 1.0 | 231.0 |
| Price inputted by respondent for last purchased meat and seafood ($) | 38.11 | 1.0 | 202.0 |
| Price inputted by respondent for last purchased dairy and eggs ($) | 31.96 | 1.0 | 130.0 |
| Price inputted by respondent for last purchased other food ($) | 32.79 | 1.0 | 110.0 |

In conclusion, this section presents a detailed case study and data collection framework aimed at investigating the demand for food products in the U.S. The study's systematic approach thoroughly examines the diverse factors influencing consumer choices for these products. Significant variations in food supply chain awareness categories highlight shifting consumer perspectives driven by external factors. Additionally, the dataset includes detailed price distributions across different food categories, providing valuable insights into consumer preferences and behaviors. The next section will delve into how this dataset is used within a machine learning modeling framework to present its findings.

*4.4. Machine Learning Modeling*

Training and validation of the different machine learning models are performed locally in an Apple Macbook Pro laptop with Apple M2 Pro chip, 10-core CPU, 16-core GPU, 16-core Neural Engine, 16GB unified memory and 512GB SSD storage, with a macOS Sonoma 14.4.1 system using the Python 3.8.0 programming language, which allows the access of open-source libraries like pandas, Scikit-learn, PyCaret, SHAP, and etc. These libraries empower various functionalities such as data processing, training machine learning models, optimizing hyperparameters, conducting validation, and evaluating SHAP. (Ali, 2020; Ke et al., 2017a; S. Lundberg & Lee, 2017; McKinney, 2010; Pedregosa et al., 2011).

The constructed dataset undergoes typical steps employed in data mining processes. These steps include removing rows with multiple null values, eliminating outliers, normalizing numerical features to scale their values, and converting categorical features into dummy variables to enhance model performance. (Pineda-Jaramillo & Arbeláez-Arenas, 2021).

A correlation analysis among input features is executed to eradicate redundancy in the training of machine learning models. The resultant dataset comprises 26,720 rows and 38 columns, encompassing the target variable for prediction.

Following the division of the final dataset into training and test sets, the mentioned models (decision tree classifier, extra trees classifier, random forest classifier, LightGBM, and multinomial logit model) undergo training on the training set using the stratified k-fold cross-validation technique. Subsequently, the performance of these trained models is assessed using the AUC metric on the test set. The parameters utilized in the initial training phase correspond to the default settings of the models within open-source Python libraries, with detailed documentation available for further reference. (Ali, 2020; Ke et al., 2017a; S. Lundberg & Lee, 2017; McKinney, 2010; Pedregosa et al., 2011).

Table 3 displays the outcomes of training the initial models. Notably, the LightGBM model demonstrates superior AUC performance for both the training and test sets compared to other models. Consequently, the LightGBM model progresses to the subsequent phase for hyperparameter optimization, resulting in an enhanced AUC score showcased. Previous research suggests that well-calibrated models typically exhibit an AUC exceeding 0.60. (Alshomrani et al., 2015; Calle et al., 2011; Wald et al., 2013).

Consequently, the optimized LightGBM model is chosen to assess the influence of class label attributes within each class. LightGBM is a tailored implementation of gradient boosting, an ensemble technique that constructs numerous weak-learner models to compose a robust learner, with each weak-learner model achieving high precision for a subset of the dataset. Developed by Microsoft, LightGBM is an open-source library designed for training gradient boosting models with enhanced efficiency. Noteworthy advantages include accelerated training speed, heightened efficiency and precision, minimized memory usage, among others.(Ke et al., 2017a; Pineda-Jaramillo, 2021).

**Table 3** Results of the machine learning models.

| Model | AUC in training set | AUC in test set |
| --- | --- | --- |
| Light Gradient Boosting Machine (LightGBM) | 0.692 | 0.672 |
| Gradient Boosting Model | 0.651 | 0.645 |
| Random Forest Classifier | 0.643 | 0.629 |
| Extra Trees Classifier | 0.629 | 0.619 |
| K Neighbors Classifier | 0.604 | 0.577 |
| Decision Tree Classifier | 0.574 | 0.569 |
| Optimized LightGBM model | 0.701 | 0.699 |

## 5. Results

This section unveils results from a meticulous analysis employing a machine learning model. It explores the factors shaping consumer choices for food products with transportation certificates, delving into attributes' influence on preferences, highlighting price sensitivity, certificate significance, transportation, and sociodemographic impacts.

Following multiple iterations and a comprehensive assessment of potential independent variables, the LGBM model demonstrates a higher desirability of products with transportation certificates compared to those lacking transportation certificates, with the least preferred option being the decision not to make a purchase.

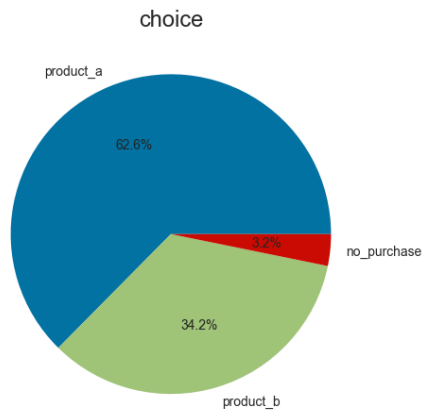

**Figure 9**. Respondent's Choice Distribution

**Figure 9** shows the respondent's choice upon three alternatives. Product A ranks as the top choice, followed by Product B as the second preferred option, while No Purchase emerges as the least favored choice.

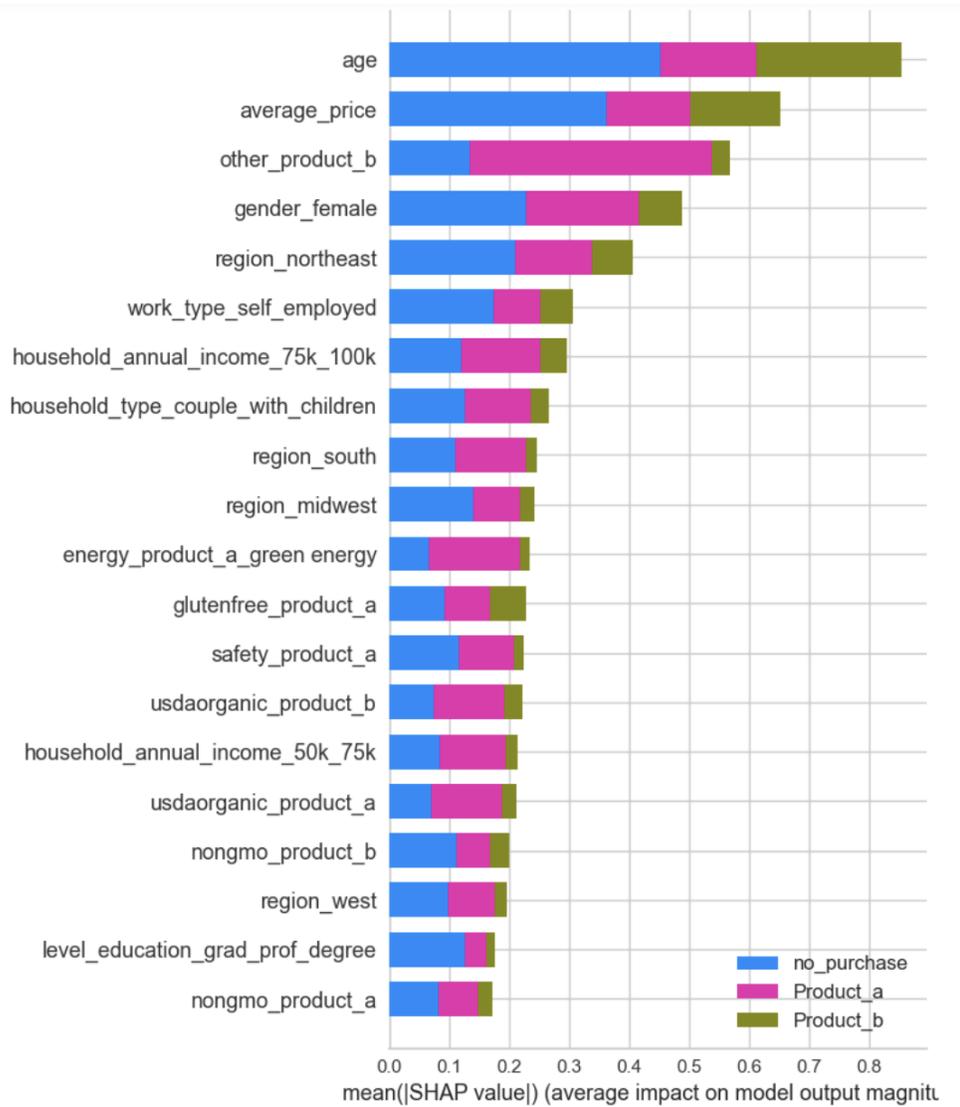

**Figure 10**. SHAP plot with the global class-label-attribute importance in the optimized LightGBM model.

**Figure 10** illustrates the effects of class label attributes on the respective class labels, showcasing the direct influence of each attribute on the outcomes of the optimized LightGBM model.

**Figure 10** arranges the class label attributes based on their overall impact, as determined by the SHAP values. Attributes appearing higher in the ranking within **Figure 10** signify a more significant influence on their respective class labels. Moreover, **Figure 10** aids in comprehending how each class label attribute contributes to the selection of its corresponding class label.

In the following subsections, we methodically dissect the multifaceted array of elements that steer consumer decisions concerning food products and transportation certificates. We investigate the role of product-specific factors, such as price dynamics and certificate relevance, in shaping consumer choices. Additionally, we meticulously examine transportation certificate factors, including transportation mode, energy source during transportation, safety, IoT, and MABDs certificates. Furthermore, a comprehensive analysis of socio-demographic variables, encompassing age, gender, employment type, education level, household income, household type, and regional location, is presented. These insights provide a substantive understanding of consumer behavior, which can guide enhancements in the food supply chain and marketing strategies.

*5.1. Product-Specific Factors*

Within this subsection, we delve into the intricate interplay of product-specific factors that profoundly shape consumer choices. We explore how variations in price sensitivity and the influence of certificates, such as USDA Organic, Non-GMO, Gluten-free, and others, impact consumer preferences.

Average price is the 2nd most critical factor in **Figure 10**. The results indicate that with price increases, respondents tend to not make a purchase, which is consistent with the first survey. Consumers carefully consider prices before making purchasing decisions. When the price is high, they are more inclined to not purchase.

Other certificate of Product B is the most important factor when choosing a food product alternative as it ranks the 3rd place in **Figure 10**. When product B has other certificate available, consumers prefer to choose product A and very unlikely to choose product B. It is understandable that other certificates other than USDA, Gluten-free, and Non-GMO, may not sound familiar to consumers, which lead to not choosing the product.

The presence of a gluten-free certificate for Product A (ranked 12th) tends to discourage people from choosing to make a purchase. Gluten-free food is gaining popularity due to its perceived health benefits by those that are not gluten intolerant – whether that is true or not. Interestingly, transportation certificate may substitute the effect of gluten-free certificates for such gluten-tolerant demand.

When product A and product B both have USDA Organic certificates (ranked 16th and 14th, respectively), users still prefer product A. This finding supports initial survey's result as there is a strong inclination toward organic food and even stronger with transportation certificates.

Even with both product A and product B having Non-GMO certificates (ranked 17th and 20th, respectively), users still lean towards not making a purchase. Non-GMO foods are becoming increasingly popular because they are often perceived as healthier, but the results indicate a lack of interest among consumers in purchasing food products with non-GMO certificates.

*5.2. Transportation Certificate Factors*

This subsection explores different certificates of transportation, including transportation mode, energy source during transportation, safety, IoT, and MABDs certificate. It reveals noteworthy consumer preferences associated with these transportation related factors. The results emphasize the requirement for further research and the potential for integrating public transportation solutions with digital food supply chain systems. This, in turn, supports the idea of pursuing more investigations and fostering collaboration between the transportation and supply chain sectors,

Product A with green energy (ranked 11th) certificate tends to be favored by more than half of the respondents. It confirmed our assumption that nowadays people are more concerned about the environment and would like to pay for eco-friendly products. Emission during food transportation accounts for 28% of the total energy use in the U.S. In a single year[reference]. Cleaner energy is necessary for the environment and the consumer preferences.

The safety certificate associated with Product A (ranked 13th) appears to deter people from purchasing the food product. This could stem from the increasing occurrence of food safety issues, leading to a sense of distrust towards safety certificates. However, those who place trust in safety certificates are still inclined to purchase Product A.

It is essential to highlight that certain transportation certificates, i.e. energy source and safety certificates, displayed statistical significance in consumer food product choices. Conversely, factors like transportation mode, IoT, and MABDs show no significant impact among respondents. The observed trends in consumer preferences regarding food purchases may stem from differing priorities. The growing interest in transportation traceability likely arises from concerns about product safety and quality assurance. Opting for safety certificates might indicate a preference for transparent production processes. Conversely, the lack of attention to transportation mode, IoT, and MABDs could be due to their perceived minimal effect on product quality or safety. These trends underscore how consumers thoughtfully evaluate information to align with their priorities when selecting food items, underscoring the significance of transparency and safety.

*5.3. Decision-maker Factors*

This section outlines how decision-maker factors impact food choices, with a focus on age, gender, employment type, education level, household income, household type, and regional location. The following discussion provides in-depth insights into these influences.

Age ranks 1st place in the importance chart **Figure 10**. Although many aging users opt not to make a purchase, the rest divide between product A and product B, with a higher preference towards product B among the respondents. This outcome mirrors our findings from the first survey, showing that as age rises, respondents are increasingly inclined to not make a purchase, which suggests that with advancing age, customers tend to become less adaptable to changes in products.

Gender plays an important role in selecting food products. Female (ranked 4th) respondents have a preference on product A over product B, but around 1/2 female demand will select not purchase at all. This observation could be linked to the tendency for females to exhibit lower levels of risk-taking behavior compared to males ("National Safety Council, Odds of Dying, Injury Facts.," accessed April 6, 2024).

People in the Northeast (ranked 5th), Midwest (ranked 10th) and West (ranked 18th) of the U.S. are more inclined to not purchase while these in the South (ranked 9th) tend to have a preference on product A. However, across all four regions, the likelihood of choosing product B is low. This implies that consumers in the northeast U.S. tend to approach their purchase decisions for food products with a more discerning mindset.

Self-employed worker (ranked 6th) will mostly choose no purchase option as shown in **Figure 10**. Less than 1/2 of the respondents choose either product a or product B, which indicates that self-employed workers are highly cost-conscious and may consider products with or without transportation certificates as higher-priced options, which diminishes their attractiveness.

Household annual income is ranked 7th (75K-10K) and 15th (50K-75K) in **Figure 10**. When household annual income falls within the 75K-10K range, nearly an equal percentage of people opt for either no purchase or product A, with a slightly higher proportion favoring product A. For consumers with 50K-75K household income, product A is a preferred option over no purchase, followed by product B. This outcome reveals that most households with middle level annual income value the presence of transportation certificates.

Most couple respondents living in household with children (ranked 8th) will select not to purchase. Consumers with kids may be more sensitive to food safety issues, which prevent them to purchase food product with unfamiliar certificates. Interestingly, just lightly less respondents chose product A, which reveals that they are willing to trust the product with more certificates.

Respondents with an education level of graduate or professional degree (ranked 19th) tend to choose not to purchase. Highly educated consumers might be more critical and discerning, which makes them perceive certificate labels as marketing tactics rather than indicators of genuine quality or safety. They might possess a deeper understanding of food production methods and the limitations of certain certifications, which could influence their decision-making process.

While other variables do exert some influence on the selection of food products, their SHAP values are too low to prompt significant changes in choice, and thus, they are not addressed.

## 6. Discussion

The analysis reveals several key insights regarding factors influencing food product selection.

Age emerges as the most critical factor, with older individuals showing a tendency to opt out of purchases, particularly as their age increases, suggesting a reduced adaptability to product changes. Gender also plays a role, with females exhibiting a preference for non-purchase. Regional differences highlight distinct purchasing behaviors, with consumers in the Northeast, Midwest, and West more inclined towards non-purchase decisions while consumers from the South prefer Product A. Self-employed individuals typically opt for non-purchase options, indicating a high degree of cost-consciousness. Household income levels influence preferences, with transportation certificates being valued among middle-income households. Highly educated consumers tend to lean towards non-purchase decisions, possibly due to a critical appraisal of certificate labels.

Safety concerns influence choices, with safety certificates potentially deterring purchases. Environmental considerations, such as green energy certificates, drive preferences towards eco-friendly Product A.

Price is the second most important consideration, with higher prices often leading to non-purchase decisions. The presence of alternative certificates, particularly for Product B, significantly influences choices, with consumers favoring Product A when transportation certificates are available. USDA certificate serves as an additional incentive for individuals to lean towards selecting Product A over alternative options while conversely, the presence of Gluten-Free and Non-GMO certificates tend to dissuade individuals from making a purchase.

Overall, these findings underscore the multifaceted nature of consumer decision-making in the food product domain, influenced by demographic factors, price sensitivity, certificate perspective, safety considerations, and environmental consciousness.

## 7. Conclusion

This study represents the inaugural investigation into the U.S. market's interest in food products accompanied by transportation certificates. Employing a machine learning model, we utilize Stated Preference data to shed light on consumer preferences between products with and without transportation certificates. Our research is spurred by the increasing awareness among Americans concerning food supply chains and the capacity of transportation elements to address health considerations, improve quality of life, and promote fairness.

The results reveal a clear preference in the U.S. market for products accompanied by transportation certificates compared to those without, with abstaining from purchase emerging as the least preferred option. Particularly, consumers prioritize safety and energy source certificates as essential transportation attributes. In contrast, the impact of transportation mode, IoT, and MABDs on the U.S. market is statistically negligible.

Regarding product characteristics, pricing negatively influences purchasing decisions. Moreover, several food certificates enhance the value of purchases. Among them, other certificate stands out as the most influential across all purchase categories. However, for food items lacking transportation certificates, Gluten-free and non-GMO certificates hold significance, while USDA certificates are primarily relevant when a transportation certificate is included.

When considering decision-making factors, individuals with middle-level household incomes and those residing in the South region tend to favor purchasing products with transportation certificates. Conversely, older, female, self-employed, highly educated consumers, as well as individuals from the Northeast, Midwest, and West regions, along with those from families with children, are more inclined to refrain from making a purchase.

However, this study also has some limitations. Since our online survey was distributed within the United States, data collection was confined to this domestic context. Furthermore, the study focused solely on analyzing the three most prevalent certificates in the market—USDA, Non-GMO, and Gluten-Free—resulting in limitations to the scope of the research. Besides, the significance of other certificates has been revealed in the survey, indicating the necessity to further explore additional food certifications. Moreover, transportation mode, IoT and MABD certificates did not turn out to be significant, which encourages us to explore other parts of transportation.

Future research can explore ways in which transportation agencies can synergize V2X solutions with digital engineering developments coming from the private sector to support the transportation traceability desired by Americans. Data from other regions could be collected to study the global demand for FSCT. Certificates other than the ones used in the survey could be considered to itemize their impacts (i.e., Certified Vegan, Keto, Whole30, and BAP). Other transportation aspects can be explored, such as ethical, which involves in fair labor practices, animal welfare during transportation, and adherence to regulatory standards. Moreover, it would be interesting to create sensitivity analysis to link the findings of the tuned model trained in this study to the wider U.S. population. This could involve using a modeling method to forecast the specific food choices made by individual Americans.

## 8. Declaration of generative AI and AI-assisted technologies in the writing process

During the preparation of this work the authors used GPT-3.5 in order to improve language and readability. After using this tool/service, the author(s) reviewed and edited the content as needed and take full responsibility for the content of the publication.